\documentclass[conference]{IEEEtran}
\IEEEoverridecommandlockouts
\usepackage{cite}

\usepackage{amsmath,amssymb,amsfonts,mathtools,bm}
\usepackage{amsthm}
\usepackage{algorithm}
\usepackage{algorithmic}
\usepackage{graphicx}
\usepackage{textcomp}
\usepackage{xcolor,float}
\usepackage{tabularx}
\usepackage{textcomp}
\usepackage{diagbox}
\usepackage{svg}
\usepackage{booktabs}
\usepackage{subfigure}
\usepackage{hyperref}

\allowdisplaybreaks
\renewcommand{\baselinestretch}{1}

\begin{document}

\title{Distilling Knowledge from Resource Management Algorithms to Neural Networks: A Unified Training Assistance Approach}

\author{
\IEEEauthorblockN{
Longfei Ma\IEEEauthorrefmark{1},
Nan Cheng\IEEEauthorrefmark{1},
Xiucheng Wang\IEEEauthorrefmark{1},
Zhisheng Yin\IEEEauthorrefmark{2},
Haibo Zhou\IEEEauthorrefmark{3},
Wei Quan\IEEEauthorrefmark{4}\\
}
\IEEEauthorblockA{
\IEEEauthorrefmark{1}School of Telecommunications Engineering, Xidian University, Xi'an, 710071, China\\
\IEEEauthorrefmark{2}School of Cyber Engineering,
Xidian University, Xi'an, 710071, China\\
\IEEEauthorrefmark{3}School of Electronic Science and Engineering, Nanjing University, Nanjing, Jiangsu, 210093, China.\\
\IEEEauthorrefmark{4} School of Electronic and Information Engineering, Beijing Jiaotong University, Beijing 100044, China\\
Email: \{lfma, xcwang\_1\}@stu.xidian.edu.cn, dr.nan.cheng@ieee.org, zsyin@xidian.edu.cn,\\ haibozhou@nju.edu.cn, weiquan@bjtu.edu.cn,}}

    \maketitle

\IEEEdisplaynontitleabstractindextext

\IEEEpeerreviewmaketitle

\begin{abstract}
As a fundamental problem, numerous methods are dedicated to the optimization of signal-to-interference-plus-noise ratio (SINR), in a multi-user setting. Although traditional model-based optimization methods achieve strong performance, the high complexity raises the research of neural network (NN) based approaches to trade-off the performance and complexity. To fully leverage the high performance of traditional model-based methods and the low complexity of the NN-based method, a knowledge distillation (KD) based algorithm distillation (AD) method is proposed in this paper to improve the performance and convergence speed of the NN-based method, where traditional SINR optimization methods are employed as ``teachers" to assist the training of NNs, which are ``students", thus enhancing the performance of unsupervised and reinforcement learning techniques. This approach aims to alleviate common issues encountered in each of these training paradigms, including the infeasibility of obtaining optimal solutions as labels and overfitting in supervised learning, ensuring higher convergence performance in unsupervised learning, and improving training efficiency in reinforcement learning. Simulation results demonstrate the enhanced performance of the proposed AD-based methods compared to traditional learning methods. Remarkably, this research paves the way for the integration of traditional optimization insights and emerging NN techniques in wireless communication system optimization.
\end{abstract}


\section{Introduction}

The optimization of the signal-to-interference-plus-noise ratio (SINR) has long confounded countless researchers pursuing the evasive grail of the state-of-the-art (sota) algorithm. From the classical water-filling approach to cutting-edge game theory and convex optimization methods, myriads of proposed techniques promise performance improvements \cite{6805125}. However, the unacceptable latency in implementing these complex algorithms impedes their practical application. Thus the meteoric rise of efficient deep learning methods has opened new prospects, spurring the exploration of neural network (NN)-based techniques to optimize SINR \cite{8743390}. The flexibility of NNs - from multilayer perceptrons (MLP) \cite{8444648} to graph neural networks (GNN) \cite{shen2020graph,wang2023scalable,yang2023knowledge}, combined with varied training techniques like reinforcement learning \cite{cheng2019space} and unsupervised learning \cite{shen2020graph} - suggests that with sufficient data, even designers unfamiliar with communications theory can develop satisfactory SINR optimization. This begs the rethinking of the question: \textit{what is the essence of traditional optimization versus neural techniques}? Clarifying their distinct features can guide superior algorithm design.

Traditional methods boast outstanding performance given ample computational resources, even surpassing sophisticated NN-based techniques including GNN \cite{shen2020graph}. In contrast, NNs offer comparable performance through simple matrix multiplications and low latency \cite{goodfellow2016deep,wang2022joint}, sacrificing the traditional optimization for blazing speed. Besides performance and inferencing latency, training NNs presents unique obstacles in optimizing SINR. While high performance requires expansive datasets and extended training, simply extracting network features fails to capture a vital characteristic: the optimality guarantees of classical techniques. Rather than a blank slate like Go, many SINR optimizations have known bounds, offering insights into the structure of high-quality solutions. 

Inspired by knowledge distillation (KD) \cite{hinton2015distilling}, which transfers knowledge from a large high-performance teacher NN into a small student network, the high-performance traditional optimization method is regarded as the large teacher model in this paper, where arbitrary NN models can be regarded as the student model to learning features of performance guaranteed solution from the traditional methods. The main contributions of this paper are as follows.
\begin{enumerate}
    \item An algorithm distillation (AD) approach is proposed to transfer optimization knowledge from traditional SINR optimization methods to neural network (NN) models.
    \item Traditional methods are utilized as teacher models, guiding NN student models to achieve enhanced performance and accelerated training through the AD framework.
    \item The AD framework enables integrating insights from performance-guaranteed traditional optimization into diverse NN architectures (MLPs, GNNs) and training techniques (reinforcement learning, unsupervised learning).
    \item Simulation results show significant improvements in NN-optimized SINR performance and convergence speed through AD across various NN models and training approaches.
\end{enumerate}

\section{NN for SINR Optimization}
In order to exemplify that knowledge from traditional algorithms can be migrated to NNs using knowledge distillation, this paper uses the classical broadcast resource management problem as a special example to explore its auxiliary performance for neural network training.

\subsection{System Model and Problem Formulation}

\begin{figure}[ht]
  \centering
  \includegraphics[width=0.8\columnwidth]{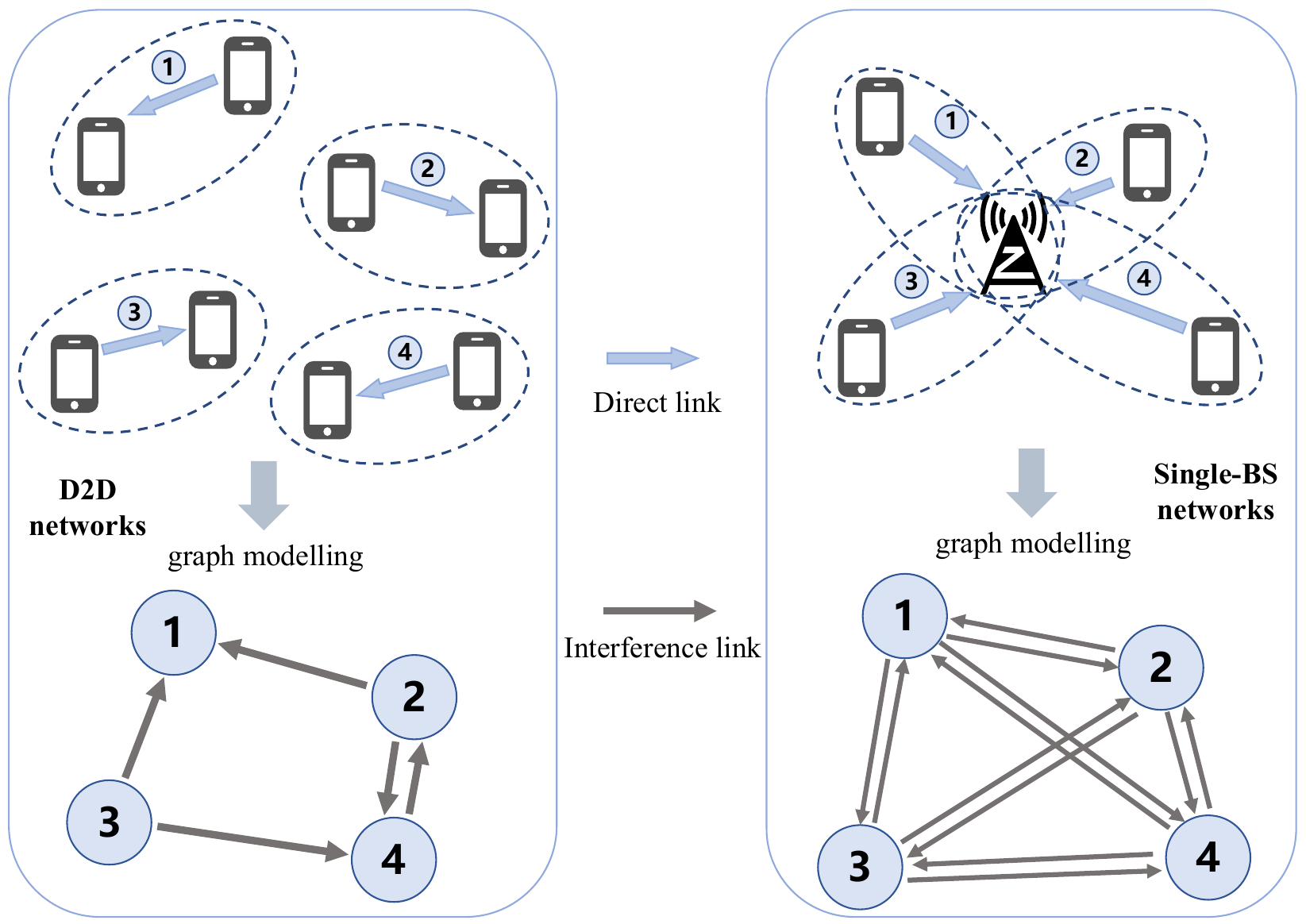}
   \vspace{-5pt}
  \centering \caption{An illustration of a $K$-user interference channel.}
  \label{model}
   \vspace{-5pt}
\end{figure}

In this paper, we consider a D2D network where there are $K$ pairs of senders and receivers who need to communicate with each other, randomly distributed in the plane, and since all users share the same frequency band for communication, there are $K\times(K-1)$ pairs of interfering links in addition to the $K$ communication links. The interference relationship between users can be modeled as a graph as shown in Fig. \ref{model}. Since the interference between distant users is small, we draw directed edges between two nodes only when the distance between the transmitter and receiver is below a certain threshold. Remarkably, this scenario setup can be easily applied to a single-BS network where multiple users use the same resource block (RB) with the same base station by simply considering the base station as the receiver in all links.

The objective of radio resource management is to maximize the sum rate of all $K$ users, which can be formulated as follows.
\begin{align}
    \max_{\bm{p}} \sum_{i=1}^{K}\log_{2}\left(1+\frac{p_{i}h_{i,i}}{\sum_{j\neq i}p_{j}h_{j,i}+\sigma^2}\right), \label{obj}
\end{align}
where $p_{i}$ is transmission power of the sender of link $i$, and $h_{i,j}$ is the channel gain from the sender in link $i$ to the receiver in link $j$.

\subsection{Different Training NN Methods for SINR Optimization}
The rapid development of NN training techniques has given rise to a number of NN training methods, including supervised learning (SUP) based on data labels, unsupervised learning (UNSUP) without data labels, and reinforcement learning (RL) by rewarding the training of NNs, and in this subsection will present how different training methods can achieve the optimization of NNs for neural networks and discuss their respective features.

$\bullet$\textbf{Supervised Learning} as an NN training method that requires the optimal solution $p^{*}$ of \eqref{obj} as the label, was widely used in early NN-based SINR optimization methods. In the process of training an NN using supervised learning, the parameters of the NN $\bm{\theta}$ need to be updated to make the broadcast resource allocation scheme $p$ of the output of the NN as similar as possible to the label $p^{*}$, so the $\bm{\theta}$ of the $\mathcal{F}$ are updated according to the following equation.
\begin{align}
    \bm{\theta} = \bm{\theta} - \alpha\nabla_{\bm{p}}\|\bm{p}^{*}-\bm{p}\|_{\bm{p}=\mathcal{F}(\bm{h},\bm{w};\bm{\theta})}\nabla_{\bm{\theta}}\mathcal{F}(\bm{h},\bm{w};\bm{\theta}),\label{sup}
\end{align}
where $\alpha$ is the learning rate. Although training NNs using supervised learning is simple and efficient, there are two insurmountable challenges to using supervised learning: obtaining optimal solutions as \textit{labels} and \textit{overfitting}. When $K$ is large, it is not possible to obtain $p^{*}$ in acceptable time using brute force search, so the so-called $p^{*}$ treated as a label is actually a solution to the traditional optimization algorithm, whose optimality cannot be guaranteed since the \eqref{obj} is non-convex, therefore, the performance of NN trained to be supervised learning is usually poorer than traditional methods. Moreover, according to \eqref{sup} the objective of the NN is to minimize the distance between the output and the label, which encourages the NN to learn the distribution of labels to minimize the $\|\bm{p}^{*}-\bm{p}\|$. Thus, once the distribution of $p^{*}$ corresponding to the test environment is different from that of the labels, the performance of the NN trained by supervised learning drops dramatically, which is also known as overfitting. However, it can also be shown that supervised learning can make the NN learn the features and distributions of the solutions corresponding to the labels, which is the inspiration for our proposed algorithm.

$\bullet$\textbf{Unsupervised Learning} is a promising method for training NNs for solving optimization problems that have emerged in the last two years. By transforming the optimization problem into a derivable loss function, the NN parameters can be updated directly using the following chain derivation.
\begin{align}
    \bm{\theta} = \bm{\theta} + \alpha\nabla_{\bm{p}}\mathcal{H}(\bm{p}|\bm{h},\bm{w})_{\bm{p}=\mathcal{F}(\bm{h},\bm{w};\bm{\theta})}\nabla_{\bm{\theta}}\mathcal{F}(\bm{h},\bm{w};\bm{\theta}),\label{unsup}
\end{align}
where $\mathcal{H}(\bm{p}|\bm{h},\bm{w})=\sum_{i=1}^{K}w_{i}\log_{2}\left(1+\frac{p_{i}h_{i,i}}{\sum_{j\neq i}p_{j}h_{j,i}+\sigma^2}\right)$. Compared to RL, using unsupervised learning for training not only also removes the dependence on labeling, enabling the NN to learn $\bm{h}$ and $\bm{w}$ features directly, but also gets rid of the delay of exploration efficiency on the training speed of the NN. However, since \eqref{obj} is non-convex, optimizing the value of $\bm{p}$ along the direction of the gradient at the point of $\bm{p}$ in the output of the NN does not guarantee its global convergence performance, although it can make $\bm{p}$ close to the local optimal point. So a feasible way to further optimize the training of unsupervised learning is to supplement the direction information between $\bm{p}$ and $\bm{p^{*}}$ when calculating the parameter gradient of NN. 

$\bullet$\textbf{Reinforcement Learning} updates the NN parameters based on the sum of the SINRs of all users that can be reached at the current $\bm{h} $and $\bm{w}$ in the $\bm{p}$ output of the NN by letting the NN interact with the environment continuously so that the $\bm{p}$ output of the NN can obtain higher values of \eqref{obj}. Since $\bm{p}$ is continual, thus, only the deterministic policy gradient (DPG) based RL algorithms that can optimize continuous variables are discussed in this paper. In the DPG, the $\bm{\theta}$ are updated as follows.
\begin{align}
    \bm{\theta} = \bm{\theta} + \alpha\nabla_{\bm{p}}\mathcal{V}(\bm{p}|\bm{h},\bm{w};\bm{\theta}_{v})\nabla_{\bm{\theta}}\mathcal{F}(\bm{h},\bm{w};\bm{\theta}),\label{rl}
\end{align}
where $\mathcal{V}$ is the value NN used to evaluate the SINR of the $\bm{p}$ output by $\mathcal{F}$ under $\bm{h}$ and $\bm{w}$. Generally, DPG-based methods are used to optimize the case where the objective function is not derivable because the objective function can be fitted using the derivable value NN $\mathcal{V}$. Despite the fact that \eqref{obj} is derivable, the use of DPG-based methods to optimize \eqref{obj} is still of interest in this paper, this is because the nature of \eqref{obj} does not dramatically affect the use of DPG-based methods, and optimizing \eqref{obj} as an example still exemplifies the DPG-based RL algorithms' ability to optimize the nature on resource management. A distinctive feature of PG can be seen in \eqref{rl}, in addition to the policy NN $\mathcal{F}$ a value NN $\mathcal{V}$ also needs to be trained, and according to \cite{silver2014deterministic} the parameters $\bm{\theta}_{v}$ of $\mathcal{V}$ are updated as $\bm{\theta}_{v} = \bm{\theta}_{v} -\alpha \|\mathcal{V}(\bm{p}|\bm{h},\bm{w};\bm{\theta}_{v})-r\|$, where $r$ is the objective value of \eqref{obj}. The performance of the $\mathcal{V}$ highly relies on the sampling efficiency. Since there are no labels in RL for the NN to learn, the NN needs to obtain higher rewards by constantly adjusting its outputs, and while this helps the NN learn outputs that perform better than the labels, it can also lead to insufficient sampling efficiency and training because the NN adopts a near-random sampling approach in exploring the performance of different outputs instead of purposely exploring the potentially superior actions. latency is high. Therefore, how to design appropriate algorithms to guide NNs to sample in the local action space where high-performance solutions are more likely to be obtained, instead of randomly exploring, will help to significantly improve the performance and training speed of RL-based NN algorithms.

\subsection{Different NN Architectures for SINR Optimization}
Different training methods can provide different parameter update directions for the NN. Still, the structural characteristics of the NN itself determine the difficulty of optimizing the NN parameters using that update direction, so a wide variety of NN architectures have been applied to optimize SINR. 

$\bullet$\textbf{MLP} is the simplest structure and the most widely used NN architecture. In the MLP-based approach, both $\bm{h}$ and $\bm{w}$ are expanded into one-dimensional vectors and concatenated together for input into the MLP, which obtains the final output after multiple cumulative matrix multiplication operations with nonlinear activation functions $\phi(\cdot)$, detailed as follows.
\begin{align}
    \mathcal{F}(\bm{h},\bm{w};\bm{\theta}) = \sum_{i} \phi(\bm{\theta}_{i}[\bm{h},\bm{w}]),
\end{align}
where $\bm{\theta}_{i}$ is trainable parameters matrix.


$\bullet$\textbf{GNN} has been the focus of more and more researchers in recent years due to its high computational efficiency and performance since it can extract topological features of communication networks. The GNN first uses the message extraction function $\varphi(\cdot)$ to obtain the features of neighbors for node $i$ as follows.
\begin{align}
    m_i =  \oplus_{j\in \mathcal{N}(i)} \varphi(h_{j,j},w_j,h_{j,i}),
\end{align}
where $\mathcal{N}(i)$ is the set of neighbors of node $i$, and $\oplus(\cdot)$ is a permutation invariant function, such as $\sum$ and $\max$. Then the transmission power $p_i$ of link $i$ is output by the node $i$ in the GNN as follows.
\begin{align}
    p_i = \psi(h_{i,i},w_{i},m_{i}).
\end{align}

Different NN architectures are used in different scenarios due to their different characteristics, and it is usually necessary to determine the kind of NN architecture to be used according to the specific scenario characteristics, so a good training assistance method should apply to a wide variety of NN architectures and can be applied in different training methods.

\section{Distilling Algorithm Knowledge Based Training Method}

\begin{figure}[ht]
  \centering
  \includegraphics[width=0.6\columnwidth]{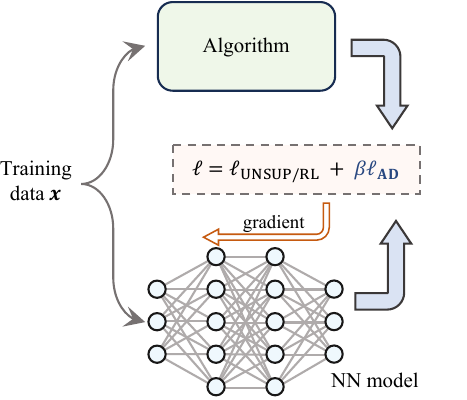}
   \vspace{-5pt}
  \centering \caption{Proposed algorithmic distillation framework, where $\bm{\ell}_{\mathrm{UNSUP/RL}}$ denotes the loss function for traditional unsupervised and reinforcement learning , and $\bm{\ell}_{\mathrm{AD}}$ denotes the supervised loss for algorithmic distillation..} 
  \label{method}
   \vspace{-10pt}
\end{figure}

\subsection{AD-Assist Unsupervised Learning and Reinforcement Learning}
Generally, the larger NN model leads to higher performance and computing complexity. However, for latency-sensitive scenarios of SINR optimization, Pursuing only high performance leads to excessive computational complexity, making the increase in decision latency of the algorithm itself difficult to compensate for the decrease in data transmission latency, which is also a general challenge in the DL areas. Therefore, in \cite{hinton2015distilling} KD is proposed to distill the knowledge from a high-performance large teacher NN $\mathcal{G}$ to a small student NN $\mathcal{F}$ by jointly minimizing the distribution divergence between the output of the teacher model and the student model and loss function value of the student model. Thus the parameters $\bm{\theta}$ of $\mathcal{F}$ are updated as follows.
\begin{align}
    &\bm{\ell} = \mathcal{H}(\bm{p})_{\bm{p}=\mathcal{F}(\cdot)} +\beta\|\bm{p}-\bm{g}\|,\\
    &\bm{\theta} = \bm{\theta} - \alpha \nabla_{\bm{p}}\,\bm{\ell} \nabla_{\bm{\theta}}\mathcal{F}(\bm{\;\cdot\;;\theta}),
\end{align}
where $\bm{p}$ and $\bm{g}$ are outputs of $\mathcal{F}$ and $\mathcal{G}$, $\bm{\ell}$ is the updating gradient of the $\bm{\theta}$, and $\beta$ is the weighting factor of KD. 
\begin{figure*}[ht]
  \centering
  \includegraphics[width=1.98\columnwidth]{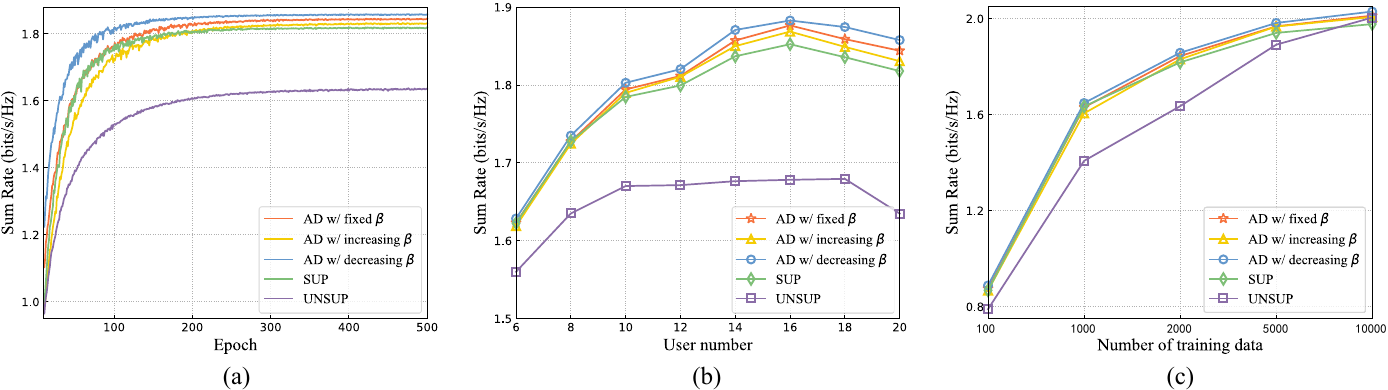}
   \vspace{-5pt}
  \centering \caption{Performance comparison of MLP training in dataset pattern.}
  \label{mlp}
   \vspace{-5pt}
\end{figure*}
Although there are no high-performance large pre-trained models in the field of broadcast resource management similar to those recognized in the fields of computer vision (CV) and natural language processing (NLP) that can be used as teacher models, the performance of some traditional methods, such as WMMSE \cite{shi2011iteratively} and FPLinQ \cite{8006944}, are recognized. Therefore, the resource allocation $\bm{p}_{fp}$ generated by the FPLinQ can be used as the teacher distribution, and the student model $\mathcal{F}$ is trained to minimize the difference between the output $\bm{p}$ and $\bm{p}_{fp}$, while obtaining the higher reward $R$ when trained by RL and minimizing the $\mathcal{H}(\bm{p}|\bm{h,w})$ when trained by unsupervised learning. 

Since unsupervised learning can directly output deterministic $\bm{p}$, in the algorithmic distillation framework, the AD term can be added directly to the chain derivation in \eqref{unsup} as 
\begin{align}
    \bm{\theta} = \bm{\theta} + \alpha\nabla_{\bm{p}}\left(\mathcal{H}(\bm{p}|\bm{h},\bm{w})+\beta\|\bm{p}-\bm{p}_{fp}\|\right)\nabla_{\bm{\theta}}\mathcal{F}(\bm{h},\bm{w};\bm{\theta}),\label{kd-unsup}
\end{align}
Similarly, DPG-based RL algorithms can be trained assisted by the AD as following 
\begin{align}
    \bm{\theta} = \bm{\theta} + \alpha(\nabla_{\bm{p}}\mathcal{V}(\bm{p}|\bm{h},\bm{w};\bm{\theta}_{v})+\beta\|\bm{p}-\bm{p}_{fp}\|)\nabla_{\bm{\theta}}\mathcal{F}(\bm{h},\bm{w};\bm{\theta}),\label{kd-rl}
\end{align}
From \eqref{kd-unsup} and \eqref{kd-rl}, the update direction of the NN gradient depends only on the output $\bm{p}$, so the AD-assisted training approach can be applied to a wide range of NN architectures, including MLPs, CNNs, GNNs, etc., since the difference lies only in the way the data features are extracted.

\subsection{Discussion of AD-Assisted Method}
\eqref{kd-unsup} and \eqref{kd-rl} can be plainly understood as the simultaneous application of supervised learning and unsupervised/RL to train the NN, which, while explaining the performance of \eqref{kd-unsup} and \eqref{kd-rl} to a certain extent, can be made possible by certain deformations to make the NN in the training process enjoy the advantages of both supervised learning and unsupervised/RL and avoid their respective disadvantages. Firstly, at the beginning of training the NN, a large $\beta$ value can be set to improve the training speed by directly learning the distribution of $\bm{p}_{fp}$ so that it no longer needs to randomly sample $\bm{p}$ in the feasible domain, and minimizing the distance of $\bm{p}_{fp}$ from $\bm{p}$ for a specific $\bm{h}$ and $\bm{w}$ can provide a gradient direction that is closer to $\bm{p}^{*}$, thus reducing the problem of the gradient direction being so different from the global optimal direction due to the non-convexity of \eqref{obj}. Moreover, by decreasing the value of $\beta$ as the number of training epochs increases, the NN also needs to learn not only the distribution of $\bm{p}_{fp}$ to reduce the error of the supervised term, but also learns how to extract features of communication networks from the $\mathcal{H}(\cdot|\bm{h},\bm{w})$  or the value $R$ of the objective function \eqref{obj} to maximize the objective value, where the overfitting challenge of supervised learning can be alleviated, which can be proved by simulation results in the next section.

\begin{figure*}[ht]
  \centering
  \includegraphics[width=1.98\columnwidth]{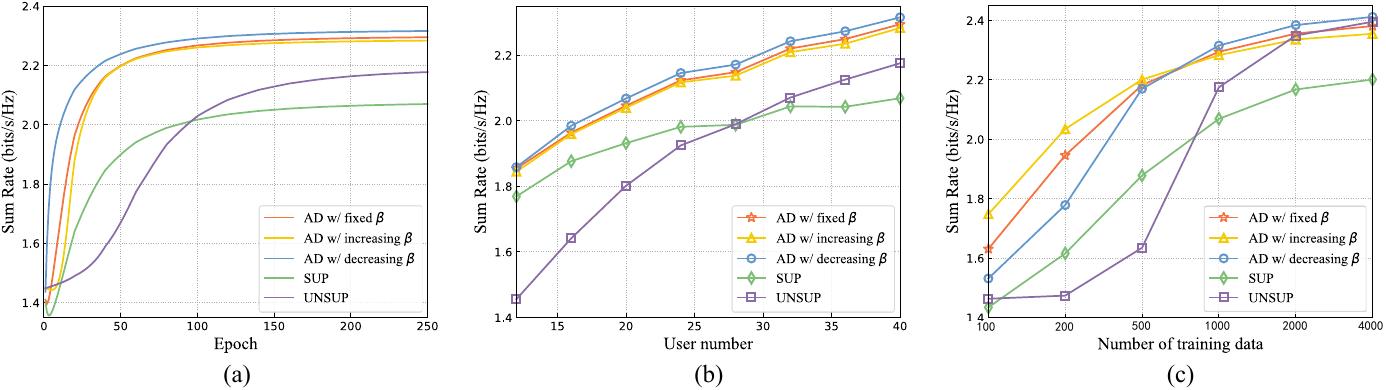}
   \vspace{-5pt}
  \centering \caption{Performance comparison of GNN training in dataset pattern.}
  \label{gnn}
   \vspace{-5pt}
\end{figure*}
\begin{figure}[ht]
  \centering
  \includegraphics[width=0.8\columnwidth]{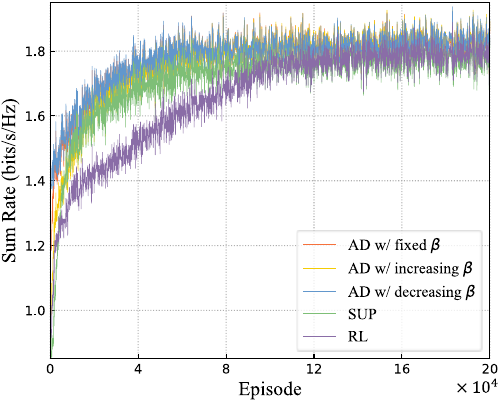}
   \vspace{-5pt}
  \centering \caption{Convergence performance of MLP in RL pattern.} 
  \label{mlprl_conver}
   \vspace{-5pt}
\end{figure}
\begin{figure}[ht]
  \centering
  \includegraphics[width=0.8\columnwidth]{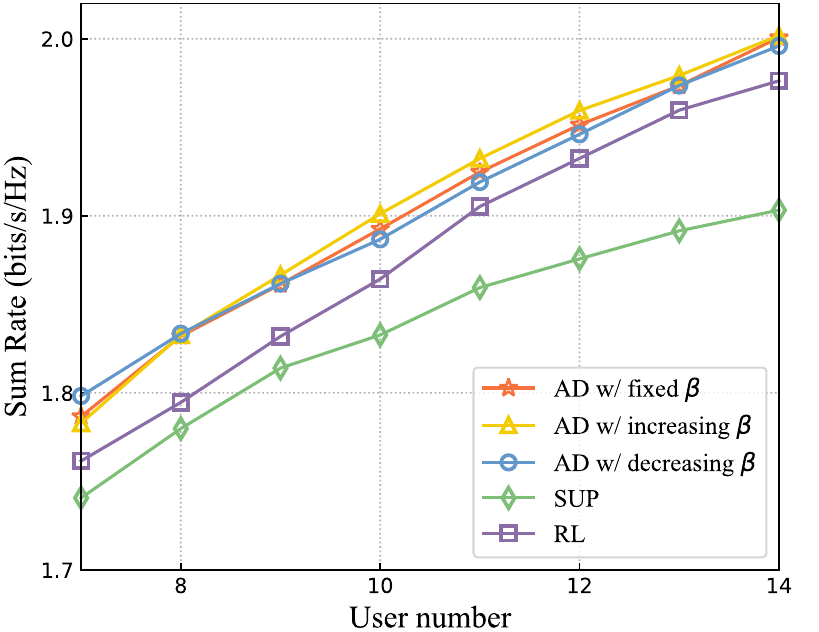}
   \vspace{-5pt}
  \centering \caption{Performance of MLP with different user numbers in RL pattern.} 
  \label{mlp_rl_usernum}
   \vspace{-10pt}
\end{figure}
\section{Simulation Results}
In this section, we provide simulation results under different neural network architectures and training patterns to validate the effectiveness of the proposed AD framework. For the proposed AD training framework, we give three strategies for setting the distillation weights $\beta$: fixed weights (AD w/ fixed $\beta$), increasing weights with training (AD w/ increasing $\beta$), and decreasing weights with training (AD w/ decreasing $\beta$). The performance of these three strategies is demonstrated in the following results. 

Based on the different sources of training data, we categorize the training patterns into two types, i.e., dataset pattern and RL pattern. Among them, dataset pattern trains the model through a fixed dataset and RL pattern trains the model by making it interact with the environment. The training objective in all cases is to maximize the sum rate. For the neural network setting, we use Pytorch to implement MLP and DGL to implement GNN \cite{8444648,9944643}. The channel state is taken as the input to the neural network and the output is the transmit power of each user. During training, we use SGD optimizer to update the model parameters.

\subsection{Performance Evaluation of AD-Assist Training in Dataset Pattern}

Fig. \ref{mlp} shows the performance of training MLP with different methods in dataset pattern. When unspecified, the number of users $K=20$, the number of training data and test data are 2000 and 1000 respectively. Fig. \ref{mlp}(a) shows the convergence performance of MLP in dataset pattern. It can be seen that in this case, ADs achieve better final performance than SUP and UNSUP, and AD w/ decreasing $\beta$ has optimal performance and convergence speed. Fig. \ref{mlp}(b) shows the performance of MLP with different user numbers in dataset pattern. It can be seen that ADs have the optimal performance. As the user number increases, the sum rate of the different methods increase first and then decrease, this is because the theoretical maximum sum rate increases as the user number increases, so the rate obtained by the model increases. However, an increase in the user number also leads to an increase in the complexity of power allocation, and when the user number increases to a certain level MLP is no longer capable of solving the problem due to the limitations of its own characterization capabilities, and therefore significantly deviates from the optimal solution, resulting in a rapid deterioration of the performance. Fig. \ref{mlp}(c) shows the performance of MLP with different numbers of training data. It can be seen that ADs achieve optimal performance in most cases. When the dataset is small, SUP performs significantly better than UNSUP, and as the data size increases, UNSUP outperforms SUP, which suggests that compared to SUP, UNSUP is more suitable for training MLP with large datasets.

Fig. \ref{gnn} shows the performance of training GNN with different methods in dataset pattern. When unspecified, the number of users $K=40$, the number of training data and test data are 1000 and 500 respectively. Fig. \ref{gnn}(a) shows the convergence performance of GNN in dataset pattern. It can be seen that ADs outperform the other methods and AD w/ decreasing $\beta$ achieves the optimal final performance and convergence speed. UNSUP has better final performance than SUP but slower convergence speed. Fig. \ref{gnn}(b) shows the performance of GNN with different user numbers in dataset pattern. It can be seen that ADs can achieve optimal performance. The performance of SUP is better when the number of users is small, while UNSUP performs better when the number of users is large. In addition, no performance deterioration with increasing number of users similar to that in Fig. \ref{mlp}(b) is observed since GNN has a stronger characterization capability than MLP in dealing with this problem. Fig. \ref{gnn}(c) shows the performance of GNN with different numbers of training data. Similar to MLP, ADs achieve optimal performance when training with different datasets. SUP outperforms UNSUP when the dataset is small, while UNSUP outperforms SUP as the number of data increases. Among the various ADs, AD w/ fixed $\beta$ and AD w/ increasing $\beta$ are more suitable for small datasets and AD w/ decreasing $\beta$ is more suitable for large datasets.

\subsection{Performance Evaluation of AD-Assist Training in RL Pattern}

Fig. \ref{mlp_rl_usernum} shows the performance of MLP with different user numbers in RL pattern. It can be seen that the ADs perform optimally and the SUP is weakest in the RL pattern. Due to the large amount of data used to train the model in RL pattern (up to several hundreds of thousands), the performance is higher than the dataset pattern for the same user number and no deterioration of the performance with the increase in the number of users is observed. Fig. \ref{mlprl_conver} shows the convergence performance of MLP in RL pattern when $K=10$. It can be seen that the final performance of ADs is better than SUP. Although the final performance of RL is close to AD, the convergence is much slower.

\begin{figure}[ht]
  \centering
  \includegraphics[width=0.8\columnwidth]{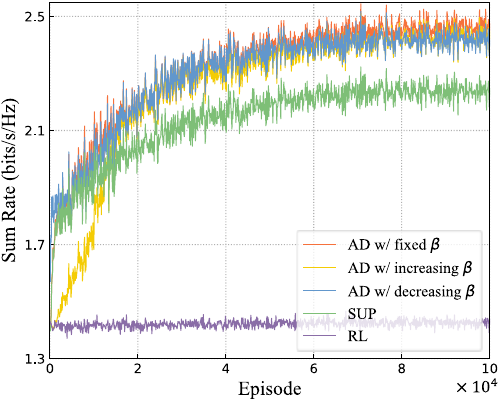}
   \vspace{-5pt}
  \centering \caption{Convergence performance of GNN in RL pattern.} 
  \label{gnnrl_conver}
   \vspace{-5pt}
\end{figure}

\begin{figure}[ht]
  \centering
  \includegraphics[width=0.8\columnwidth]{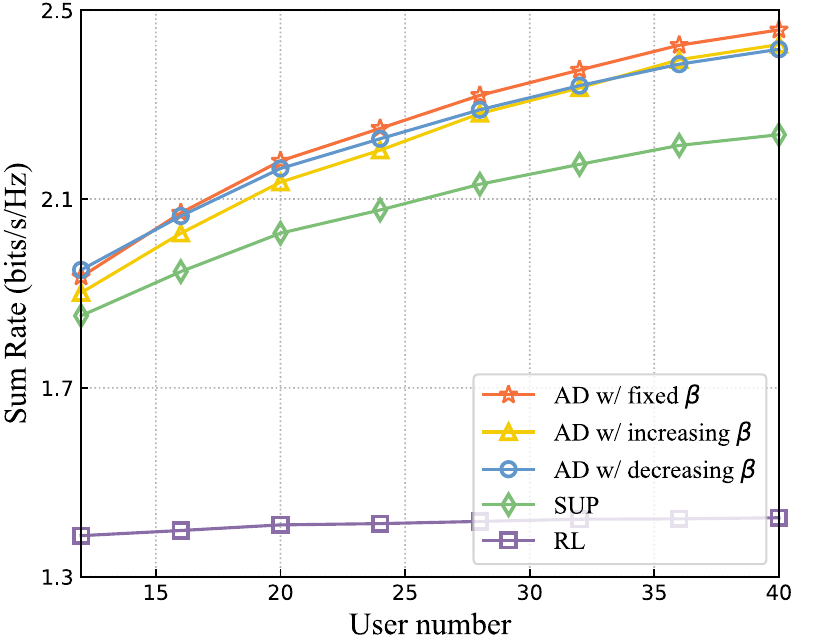}
   \vspace{-5pt}
  \centering \caption{Performance of GNN with different user numbers in RL pattern.} 
  \label{gnn_rl_usernum}
   \vspace{-5pt}
\end{figure}

Fig. \ref{gnnrl_conver} shows the convergence performance of GNN in RL pattern when $K=40$. It can be seen that ADs and SUP converge at a similar speed and the final performance of ADs is significantly better than that of SUP. Furthermore, the RL method fails to converge at all, suggesting that the RL method is not applicable to GNN architectures in this problem.
Fig. \ref{gnn_rl_usernum} shows the performance of GNN with different user numbers in RL pattern. It can be seen that similar to MLP, as the number of users increases, the performance of ADs and SUP improves with it, with ADs always outperforming SUP. Since RL cannot converge in this scenario, it always maintains poor performance.

\section{Conclusion}
In this paper, an AD-based NN training assistance method is proposed for SINR optimization in wireless networks was conducted, which efficiently mitigated the limitations inherent in the existing NN training methods, including unsupervised, and RL. Simulation results have proven the proposed AD approach significantly can enhance the learning efficiency and performance of the existing methods by utilizing the targets provided by traditional algorithms. By applying the proposed scheme in the network, the performance and convergence speed of numerous NN-based optimization methods can be enhanced by training assisted with proposed AD methods. For further research, we will explore how to effectively utilize the knowledge of traditional algorithms to improve training results in more complex problems and NN training methods.

\bibliography{ref}
\bibliographystyle{IEEEtran}

\ifCLASSOPTIONcaptionsoff
  \newpage
\fi

\end{document}